\documentclass[10pt,twocolumn,letterpaper]{article}

\usepackage{cvpr}
\usepackage{times}
\usepackage{epsfig}
\usepackage{graphicx}
\usepackage{amsmath}
\usepackage{amssymb}
\usepackage{color}

\usepackage[pagebackref=true,breaklinks=true,letterpaper=true,colorlinks,bookmarks=false]{hyperref}

\cvprfinalcopy 

\def\cvprPaperID{*}

\newcommand{\smallparagraph}[1]{\smallskip \noindent \textbf{#1}}

\ifcvprfinal\fi
\begin{document}

\title{Dynamic Deep Neural Networks: \\ Optimizing Accuracy-Efficiency Trade-offs by Selective Execution}

\author{
Lanlan Liu \qquad Jia Deng\\
{\tt\small llanlan@umich.edu \qquad jiadeng@umich.edu} \\
University of Michigan\\
2260 Hayward St, Ann Arbor, MI, 48105, USA 
}
\maketitle

\begin{abstract}
We introduce Dynamic Deep Neural Networks (D$^2$NN), a new type of feed-forward deep neural 
network that allows selective execution. Given an input, only a subset of D$^2$NN neurons
are executed, and the particular subset is determined by the D$^2$NN itself.
By pruning unnecessary computation depending on input, D$^2$NNs
provide a way to improve computational efficiency. To achieve dynamic selective execution,
a D$^2$NN augments a feed-forward deep neural network (directed acyclic graph of
differentiable modules) with controller modules. Each controller module is
a sub-network whose output is a decision that controls whether other modules can execute.
A D$^2$NN is trained end to end. Both regular and controller modules in a D$^2$NN are
learnable and are jointly trained to optimize both accuracy and efficiency. Such training
is achieved by integrating backpropagation with reinforcement learning. With extensive
experiments of various D$^2$NN architectures on image classification tasks, we demonstrate that D$^2$NNs are general and flexible, and
can effectively optimize accuracy-efficiency trade-offs. 
\end{abstract}

\section{Introduction}

This paper introduces Dynamic Deep Neural Networks (D$^2$NN), a new type of feed-forward
deep neural network (DNN) that allows selective execution. That is, given an input, only a
subset of neurons are executed, and the particular subset is determined by the network
itself based on the particular input. In other words, the amount of computation
and computation sequence are dynamic based on input. This is different from
standard feed-forward networks that always execute the same computation sequence
regardless of input. 

\begin{figure*}[t] \centering 
\includegraphics[width=0.95\textwidth]{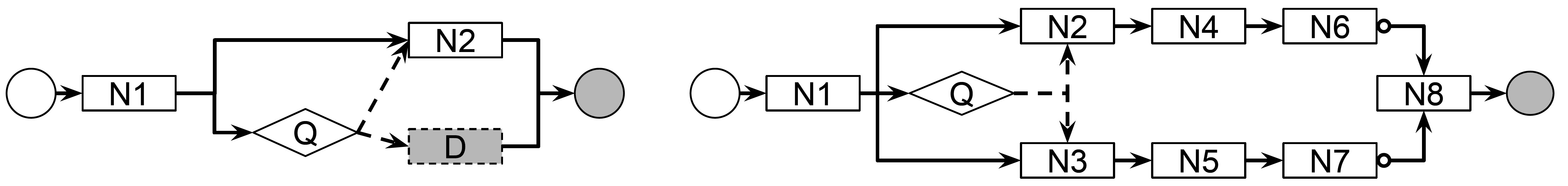}
\caption{Two D$^2$NN examples. Input and
output nodes are drawn as circles with the output nodes shaded. Function nodes are
drawn as rectangles (regular nodes) or
diamonds (control nodes). Dummy nodes are shaded.
Data edges are drawn as solid arrows and control edges as
dashed arrows. A data edge with a user defined default value is decorated with a circle.}
\label{fig:d2nn_example}
\end{figure*}

A D$^2$NN is a feed-forward deep neural network (directed acyclic graph of
differentiable modules) augmented with one or more control modules. A control module is a
sub-network whose output is a decision that controls whether other modules can
execute. Fig.~\ref{fig:d2nn_example} (left) illustrates a simple D$^2$NN with one control module (Q) and two
regular modules (N1, N2), where the controller Q outputs a binary decision on whether
module N2 executes. For certain inputs, the controller may decide that N2
is unnecessary and instead execute a dummy node D to save on computation.
As an example application, this D$^2$NN can be used for binary classification of
images, where some images can be rapidly classified as negative after only a small amount of
computation.

D$^2$NNs are motivated by the need for computational efficiency, in particular, by the
need to deploy deep networks on mobile devices and data centers. Mobile devices are
constrained by energy and power, limiting the amount of computation that can be executed. 
Data centers need energy efficiency to scale to higher
throughput and to save operating cost. D$^2$NNs provide a way to improve computational
efficiency by selective execution, pruning unnecessary computation depending on
input. D$^2$NNs also make it possible to use a bigger network under a
computation budget by executing only a subset of the neurons each time.

A D$^2$NN is trained end to end. That is, regular modules and control modules are
jointly trained to optimize both accuracy and efficiency.  We achieve such training by
integrating backpropagation with reinforcement learning, necessitated by the
non-differentiability of control modules. 

Compared to prior work that optimizes computational efficiency in computer vision and
machine learning, our work is distinctive in four aspects: (1) the decisions
on selective execution are part of the network inference and are learned end to
end together with the rest of the network, as opposed to hand-designed or separately
learned \cite{li2015convolutional,sun2013deep,DBLP:conf/icml/AlmahairiBCZLC16};
(2) D$^2$NNs allow more flexible network architectures and execution sequences
including parallel paths, as opposed to architectures with less variance~\cite{denoyer2014deep,shazeer2017outrageously}; 
(3) our D$^2$NNs directly optimize arbitrary efficiency metric that is defined by the user, while previous work has no
such flexibility because they improve efficiency indirectly through sparsity constraints\cite{bengio2015conditional,bengio2013estimating,shazeer2017outrageously}.
(4) our method optimizes metrics such as the F-score that does not decompose
over individual examples. This is an issue not addressed in prior work. We will elaborate on
these differences in the Related Work section of this paper. 

We perform extensive experiments to validate our D$^2$NNs algorithms. We evaluate various D$^2$NN architectures on several tasks. They demonstrate that D$^2$NNs are general, flexible, and can
effectively improve computational efficiency. 

Our main contribution is the D$^2$NN framework that allows a user to 
augment a static 
feed-forward network with control modules to achieve dynamic selective
execution. We show that D$^2$NNs allow a wide variety of topologies while sharing a unified
training algorithm. 
To our knowledge, D$^2$NN is the first single framework that can support various qualitatively different efficient network designs, including cascade designs and coarse-to-fine designs. 
Our D$^2$NN framework thus
provides a new tool for designing and training computationally efficient neural network models. 

\section{Related work}\label{sec:relatedwork}

Input-dependent execution has been widely used in computer vision, from cascaded
detectors~\cite{viola2004robust,felzenszwalb2010cascade} to
hierarchical classification~\cite{deng2011fast,bengio2010label}. 
The key difference of our work from prior work is that we \emph{jointly} learn both visual features and control decisions
\emph{end to end},  whereas prior work either hand-designs features and control
decisions (e.g.\@ thresholding), or learns them separately. 

In the context of deep networks, two lines of prior work have attempted to improve
computational efficiency. 
 One line of work tries to eliminate redundancy in data or computation in a
way that is input-independent.
The methods include pruning networks~\cite{han2015learning,wen2016learning,alvarez2016learning}, approximating layers with simpler
functions~\cite{denton2014exploiting,zhang2016accelerating}, and using number representations of limited
precision~\cite{chen2014dadiannao,gupta2015deep}. 
The other line of work exploits the fact that not all inputs require the same amount of
computation, and explores input-dependent execution of DNNs. Our work belongs to the
second line, and we will contrast our work mainly with them. In fact, our input-dependent D$^2$NN can be combined with input-independent methods to achieve even better efficiency. 

Among methods leveraging input-dependent execution, some use pre-defined execution-control
policies. For example,  cascade methods~\cite{li2015convolutional,sun2013deep}
rely on manually-selected thresholds to control execution; Dynamic Capacity
Network~\cite{DBLP:conf/icml/AlmahairiBCZLC16} designs a way to
directly calculate a saliency map for execution control. Our D$^2$NNs, instead, are fully
learn-able; the execution-control policies of D$^2$NNs do not require manual design and
are learned together with the rest of the network. 

Our work is closely related to conditional computation methods
~\cite{bengio2015conditional,bengio2013estimating,shazeer2017outrageously}, which activate part of a network
depending on input. 
They learn
policies to encourage sparse neural activations\cite{bengio2015conditional} or sparse expert networks\cite{shazeer2017outrageously}.
Our work differs from these methods in several ways. First, our
control policies are learned to directly optimize arbitrary user-defined global performance
metrics,
 whereas conditional computation methods have only learned policies that
encourage sparsity. 
In addition,
D$^2$NNs allow more flexible control topologies. 
For example, in \cite{bengio2015conditional}, a neuron (or block of neurons) is the unit controllee of their control policies; in \cite{shazeer2017outrageously}, an expert is the unit controllee. Compared to their fixed types of controllees, our control modules can be added in any point of the network and control arbitrary subnetworks. 
Also, various policy parametrization can be used in the same D$^2$NN framework.
We show a variety of parameterizations (as different controller networks) in our D$^2$NN examples, whereas previous conditional computation works have used some fixed formats: For example, control policies are parametrized as the sigmoid or softmax of an affine
transformation of neurons or inputs~\cite{bengio2015conditional,shazeer2017outrageously}. 

Our work is also related to attention
models~\cite{denil2012learning,mnih2014recurrent,gregor2015draw}.
Note that attention
models can be categorized as \emph{hard} attention \cite{mnih2014recurrent,ba2014multiple,DBLP:conf/icml/AlmahairiBCZLC16} 
  versus \emph{soft} \cite{gregor2015draw,stollenga2014deep}. Hard
attention models only process the salient parts and discard others 
(e.g.\@ processing only a subset of image subwindows); in contrast, soft
attention models process all parts but up-weight the salient parts. 
Thus only hard
attention models perform input-dependent execution as D$^2$NNs do. 
However, hard attention models differ from D$^2$NNs because hard attention models have typically
involved only one attention module  whereas D$^2$NNs can have multiple attention
(controller) modules --- conventional hard attention models are ``single-threaded'' whereas
D$^2$NN can be ``multi-threaded''. In addition, prior work
in hard attention models have not directly optimized for accuracy-efficiency
trade-offs.
It is also worth noting that many mixture-of-experts methods~\cite{jacobs1991adaptive,jordan1994hierarchical,eigen2013learning} also involve soft attention by soft gating experts: they process all experts but only up-weight useful experts, thus saving no computation. 

D$^2$NNs also bear some similarity to Deep Sequential Neural
Networks (DSNN)~\cite{denoyer2014deep} in terms of input-dependent execution. 
However, it is important to note that although DSNNs' structures can in principle be used to optimize accuracy-efficiency trade-offs, 
DSNNs are not for the task of improving efficiency and have no learning method proposed to optimize efficiency. And the method to effectively optimize for efficiency-accuracy trade-off is non-trivial as is  shown in the following sections.
Also, DSNNs are single-threaded: it always activates exactly one path in the computation graph, whereas for D$^2$NNs it is
possible to have multiple paths or even the entire graph activated. 

\section{Definition and Semantics of D$^2$NNs}

Here we precisely define a D$^2$NN and describe its semantics, i.e.\@
how a D$^2$NN performs inference.

\smallparagraph{D$^2$NN definition} 
A D$^2$NN is defined as a directed acyclic graph (DAG) without duplicated edges. Each node can be one of the
three types: input nodes, output nodes, and function nodes. An input or output
node represents an input or output of the network (e.g. a vector). A function node
represents a (differentiable) function
that maps a vector to another vector. Each edge can be one of the two types: data edges and control
edges. A data edge represents a vector sent from one node to another, the same as in a
conventional DNN. A control edge represents a control signal, a scalar, sent from
one node to another. A data edge can optionally
have a user-defined ``default value'', representing the output that will still be sent even
if the function node does not execute.

For simplicity, we have a few restrictions on valid D$^2$NNs: (1) the outgoing edges from a node are either all data
edges or all control edges (i.e.\@ cannot be a mix of data edges and control edges); (2) if a
node has an incoming control edge, it cannot have an outgoing control edge. Note that
these two simplicity constraints do not in any way restrict the
expressiveness of a D$^2$NN. For example, to achieve the effect of a node with
a mix of outgoing data edges and control edges, we can just feed its data output 
to a new node with outgoing control edges and let the new node be an identity
function. 

We call a function node a \emph{control node} if its outgoing edges are
control edges. We call a function node a \emph{regular node} if its outgoing
edges are data edges. Note that it is possible for a function node to take no data input and output
a constant value. We call such nodes ``dummy'' nodes. We will see that the ``default values'' and ``dummy'' nodes can significantly extend the flexibility of D$^2$NNs.
Hereafter we may
also call  function nodes  ``subnetwork'', or ``modules'' and will use these terms
interchangeably. 
Fig.~\ref{fig:d2nn_example} illustrates simple D$^2$NNs with all kinds of nodes and edges.

\smallparagraph{D$^2$NN Semantics} 
Given a D$^2$NN, we perform inference by traversing the graph
starting from the input nodes. Because a D$^2$NN is a DAG, we can execute each node in a
topological order (the parents of a node are ordered before it; we take both data edges and control edges in consideration), same as
conventional DNNs except that the control nodes can cause the computation
of some nodes to be skipped. 

After we execute a control node, it outputs a set of control scores, one for each of its
outgoing control edges. The control edge with the highest score is ``activated'', meaning that the
node being controlled is allowed to execute. The rest of the control edges are not activated,
and their controllees are not allowed to execute. For example, in
Fig~\ref{fig:d2nn_example} (right), the node Q controls N2 and N3. Either N2 or N3 will execute
depending on which has the higher control score.  

Although the main idea of the inference (skipping nodes) seems simple, due to D$^2$NNs' flexibility, the inference topology can
be far more complicated. For example, in the case of a
node with multiple incoming control edges (i.e.\@ controlled by multiple controllers), it
should execute if any of the control edges are activated. 
Also, when the execution of a node is skipped, its output will be either the default value or null. If the output is the default value, subsequent execution will
continue as usual. If the output is null, any
downstream nodes that depend on this output will in turn skip execution and have a null 
output unless a default value has been set. This ``null'' effect will propagate to the
rest of the graph. Fig.~\ref{fig:d2nn_example} (right) shows a slightly more complicated example with default values: if N2 skips execution and
outputs null, so will N4 and N6. But N8 will execute regardless because its input data edge has
a default value. In our Experiments Section, we will demonstrate more sophisticated D$^2$NNs.

We can summarize the semantics of D$^2$NNs as follows: a D$^2$NN executes the same way as a conventional
DNN except that there are control edges that can cause some nodes to be skipped. A control
edge is active if and only if it has the highest score among all outgoing control edges from a
node. A node is skipped if it has incoming control edges and none of them is
active, or if one of its inputs is null. If a node is skipped, its output will be either
null or a user-defined default value. A null will cause downstream nodes to be skipped whereas a default
value will not. 

A D$^2$NN can also be thought of as a program with conditional statements. Each data edge
is equivalent to a variable that is initialized to either a default value or null. Executing a function
node is equivalent to executing a command assigning the output of the function to the
variable. A control edge is equivalent to a boolean variable initialized to False. A
control node is equivalent to a ``switch-case'' statement that computes a score for each of
the boolean variables and sets the one with the largest score to True. Checking the
conditions to determine whether to execute a function is equivalent to enclosing the
function with an ``if-then'' statement. A conventional DNN is a program with
only function calls and variable assignments without any conditional statements, whereas a
D$^2$NN introduces conditional statements with the conditions themselves generated by
learnable functions. 

\section{D$^2$NN Learning} 

Due to the control nodes, a D$^2$NN cannot be trained the same way as a conventional
DNN. The output of the network cannot be expressed as a differentiable
function of all trainable parameters, especially those in the control nodes. As a
result, backpropagation cannot be directly applied.
The main difficulty lies in the control nodes, whose outputs are discretized into control
decisions. This is similar to the situation with hard attention models~\cite{mnih2014recurrent,ba2014multiple},
which use reinforcement learning. Here we adopt the same general strategy. 

\smallparagraph{Learning a Single Control Node} 
 For simplicity of exposition we start with a special case where there is only
one control node. We further assume that all parameters except those of this control node
have been learned and fixed. That is, the goal is to learn the parameters of the control
node to maximize a user-defined reward, which in our case is a combination of accuracy and
efficiency. This results in a classical reinforcement learning
setting: learning a control policy to take actions so as to maximize reward. 
We base our learning method on Q-learning~\cite{mnih2013playing,sutton1998reinforcement}. 
We let each outgoing control edge represent an action, and let the control node approximate
the action-value (Q) function, which is the expected return of an action given the current
state (the input to the control node).

It is worth noting that unlike many prior works that use deep reinforcement learning,
 a D$^2$NN is not recurrent. For each input to the network (e.g.\@ an image),
each control node only executes once. And the decisions of a control node
completely depend on the current input. As a result, an action taken on one input has no
effect on another input. That is, our reinforcement learning task consists of only one time step. 
Our one time-step reinforcement learning task can also be seen as a contextual bandit problem, where the context vector is the input to the control module, and the arms are the possible action outputs of the module.
The one time-step setting simplifies our Q-learning
objective to that of the following regression task: 
\begin{equation} 
L= (Q(\boldsymbol{s},\boldsymbol{a}) - r)^2, 
\label{eqn:qlearn-single}
\end{equation}
where $r$ is a user-defined reward, $\boldsymbol{a}$ is an action, $\boldsymbol{s}$ is
the input to control node, and $Q$ is computed by the control node.
As we can see, training a control node here is the same as training a
network to predict the reward for each action under an L2 loss. We use
mini-batch gradient descent; for each training example
in a mini-batch, we pick the action with the largest $Q$, execute the rest of the
network, observe a reward, and perform backpropagation using the L2 loss in Eqn.~\ref{eqn:qlearn-single}.

During training we also perform $\epsilon$-greedy exploration --- instead of always
choosing the action with the best $Q$ value, we choose a random action with probability
$\epsilon$. The hyper-parameter $\epsilon$ is initialized to $1$ and decreases over time. 
The reward $r$ is user defined. Since our goal is to optimize the trade-off between accuracy and
efficiency, in our experiments we define the reward as a combination of an accuracy metric $A$
(for example, F-score) 
and an efficiency metric $E$ (for example, the inverse of the number of multiplications),
that is, $\lambda A + (1-\lambda) E$ where $\lambda$ balances the trade-off. 

\smallparagraph{Mini-Bags for Set-Based Metrics}
Our training algorithm so far has defined the state as a single training example, 
i.e., the control node takes actions and observes rewards on each training example
independent of others. This setup, however, introduces a difficulty for optimizing for accuracy
metrics that cannot be decomposed over individual examples. 

Consider \emph{precision} in the context of binary classification. Given predictions on a
set of examples and the ground truth, precision is defined as the proportion of true
positives among the predicted positives. Although precision can be defined on a single
example, precision on a set of examples does not generally equal the average of
the precisions of individual examples. In other words, precision as a metric does not decompose
over individual examples and can only be computed using a set of examples \emph{jointly}. This is different from decomposable metrics such as
\emph{error rate}, which can be computed as the average of the error
rates of individual examples. If we use
precision as our accuracy metric, it is not clear how to define a reward independently for each
example such that maximizing this reward independently for
each example would optimize the overall precision.  
In general, for many metrics, including precision and F-score, we
cannot compute them on individual examples and average the results. Instead, we must
compute them using a set of examples as a whole.  We call such metrics
``set-based metrics''. Our learning setup so far is ill-equipped for such metrics because
a reward is defined on each example independently. 

To address this issue we generalize the definition of a state from a single input to a set
of inputs. We define such a set of inputs as a \emph{mini-bag}. With a mini-bag of images, any set-based metric can be
computed and can be used to directly define a reward. Note that a mini-bag is different from a mini-batch which is commonly used for batch updates in gradient decent methods. Actually in our training, we calculate gradients using a mini-batch of mini-bags.
Now, an action on a
mini-bag $\mathbf{s} = (s_1, \ldots, s_m)$ is now a joint action $\mathbf{a} =
(a_1,\ldots,a_m)$ consisting of individual actions $a_i$ on example $s_i$. Let
$\mathbf{Q}(\mathbf{s}, \mathbf{a})$ be the \emph{joint} action-value function on the mini-bag
$\mathbf{s}$ and the joint action $\mathbf{a}$. We constrain the parametric form of $\mathbf{Q}$
to decompose over individual examples: 
\begin{equation}
\mathbf{Q} = \sum_{i=1}^m Q(s_i,a_i), 
\label{eqn:q-decompose}
\end{equation}
where $Q(s_i,a_i)$ is a score given by the control node when choosing the action $a_i$ for
example $s_i$. We then define our new
learning objective on a mini-bag of size $m$ as 
\begin{equation}
L = (r - \mathbf{Q}(\mathbf{s},\mathbf{a})) ^2 = (r - \sum_{i=1}^m Q(s_i,a_i) )^2, 
\label{qlearn-bag}
\end{equation}
where $r$ is the reward observed by choosing the joint action $\mathbf{a}$ on mini-bag
$\mathbf{s}$.  That is, the control node predicts an action-value
for each example such that their sum approximates the reward defined on the whole
mini-bag. 

It is worth noting that the decomposition of $\mathbf{Q}$ into sums 
(Eqn.~\ref{eqn:q-decompose}) enjoys a nice property: the best joint action $\mathbf{a}^*$  under the
joint action-value $\mathbf{Q}(\mathbf{s}, \mathbf{a})$
is simply the concatenation of the best actions for individual examples because
maximizing
\begin{equation} \mathbf{a}^{*} = \arg\max_{\mathbf{a}} (\mathbf{Q}(\mathbf{s},\mathbf{a})) =
\arg\max_{\mathbf{a}} ({\sum_{i=1}^m Q(s_{i},a_{i})})
\end{equation} 
is equivalent to maximizing the individual summands:
\begin{equation} a_{i}^{*} = \arg\max_{a_{i}} Q(s_{i},a_{i}), i=1,2...m.
\end{equation} 
That is, during test time we still perform inference on each example independently.

Another implication of the mini-bag formulation is: 
 \begin{equation} 
\frac{\partial L}  {\partial x_i} =
2(r - \sum_{j=1}^mQ(s_j,a_j)) \frac{\partial Q(s_i,a_i)} {\partial x_i}, 
\label{eqn:bag-backprop}
 \end{equation}

where $x_i$ is the output of any internal neuron for example $i$ in the
mini-bag. 
This shows that there is no change to the implementation
of backpropagation except that we scale the gradient using the difference between the mini-bag
Q-value $\mathbf{Q}$ and reward $r$. 

\begin{figure*}[t] \centering
\includegraphics[width=0.95\textwidth]{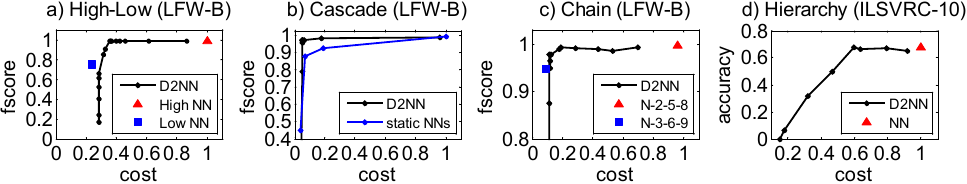}
\caption{The accuracy-cost or fscore-cost curves of various D$^2$NN architectures, as well as
  conventional DNN baselines consisting of only regular nodes. }
\label{fig:curve_4}
\end{figure*}

\begin{figure*}[t] \centering
\includegraphics[width=0.95\textwidth]{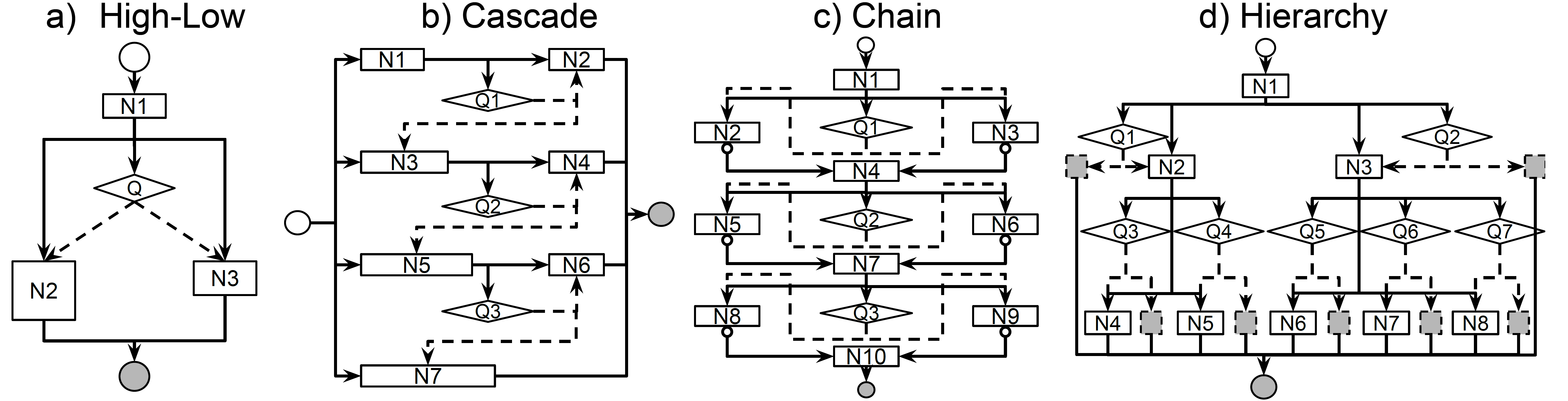}
\caption{Four different D$^2$NN architectures.}
\label{fig:d2nn_4all}
\end{figure*}

\smallskip \noindent \textbf{Joint Training of All Nodes}
We have described how to train a single control node.
We now describe how to extend this strategy to all nodes including additional control
nodes as well as regular nodes. 
If a D$^2$NN has multiple control nodes, we simply train them together. For each
mini-bag, we perform backpropagation for multiple losses together. Specifically, we
perform inference using the current parameters, observe a reward for the whole network,
and then use the same reward (which is a result of the actions of all control nodes) to
backpropagate for each control node. 

For regular nodes, we can place losses on them the same as on conventional
DNNs. And we perform backpropagation on these losses together with the control
nodes. The implementation of backpropagation is the same as conventional DNNs except that
each training example have a different network topology (execution sequence).
And if a node is skipped for a particular training
example, then the node does not have a gradient from the example.

It is worth noting that our D$^2$NN framework allows arbitrary losses to be used for
regular nodes. For example, for classification we can use the
cross-entropy loss on a regular node. One important detail is that the losses on regular
nodes need to be properly weighted against the losses on the control nodes; otherwise
the regular losses may dominate, rendering the control nodes ineffective. One way to
eliminate this issue is to use Q-learning losses on regular nodes as well, i.e.\@
treating the outputs of a regular node as action-values. For example, instead of using the cross-entropy loss on the classification
scores, we treat the classification scores as action-values---an estimated reward of
each classification decision. This way Q-learning is applied to all nodes in a unified way
and no additional hyperparameters are needed to balance different kinds of losses. 
In our
experiments unless otherwise noted we adopt this unified approach. 

\section{Experiments}

We here demonstrate four D$^2$NN structures motivated by different demands of efficient network design to show its flexibility and effectiveness, and compare D$^2$NNs' ability to optimize efficiency-accuracy trade-offs with prior work.

We implement the D$^2$NN framework in Torch. Torch
provides functions to specify the subnetwork architecture inside a
function node. Our framework 
handles the high-level communication and loss propagation. 

\smallparagraph{High-Low Capacity D$^2$NN}
Our first experiment is with a simple D$^2$NN architecture that we call ``high-low
capacity D$^2$NN''. It is motivated by that we can save computation by choosing a low-capacity
subnetwork for easy examples. 
It consists of a single control nodes (Q) and three regular nodes
(N1-N3) as in Fig.~\ref{fig:d2nn_4all}a). 
The control node Q chooses between a high-capacity N2 and a
low-capacity N3; the N3 has fewer neurons and uses less
computation. The control node itself has orders of magnitude fewer computation than regular nodes (this is true for all D$^2$NNs demonstrated).

We test this hypothesis using a binary classification task in which the network
classifies an input image as face or non-face. We use the Labeled Faces in the Wild 
\cite{LFWTech,LFWTechUpdate} dataset. Specifically, we use the 13k ground truth face crops (112$\times$112
pixels) as positive examples and randomly sampled 130k background crops (with an intersection over
  union less than 0.3) as
negative examples. We hold out 11k images for validation and 22k for testing. We
refer to this dataset as LFW-B and use it as a testbed to validate the
effectiveness of our new D$^2$NN framework. 

To evaluate performace we measure accuracy using the F1 score,
a better metric than percentage of correct predictions for an unbalanced
dataset. 
We measure computational cost using the number of multiplications following prior work~\cite{DBLP:conf/icml/AlmahairiBCZLC16,shazeer2017outrageously} 
and for reproductivity. Specifically, we use the number of multiplications (control
nodes included), normalized by a conventional DNN consisting
of N1 and N2, that is, the high-capacity execution path. Note that our D$^2$NNs also allow
to use other efficiency measurement such as run-time, latency.

During training we define the Q-learning reward as a linear combination of accuracy $A$ and
efficiency $E$ (negative cost): $r = \lambda A + (1-\lambda) E$ where
$\lambda\in [0,1]$. We train instances of high-low capacity D$^2$NNs using different
$\lambda$'s. As $\lambda$ increases, the learned D$^2$NN trades off efficiency for
accuracy. Fig.~\ref{fig:curve_4}a) plots the accuracy-cost curve on the test
set; it also plots the accuracy and efficiency achieved by a conventional DNN with only
the high capacity path N1+N2 (High NN) and a conventional DNN with only the low capacity path
N1+N3 (Low NN). 

\begin{figure*}[t] \centering
\includegraphics[width=0.9\textwidth]{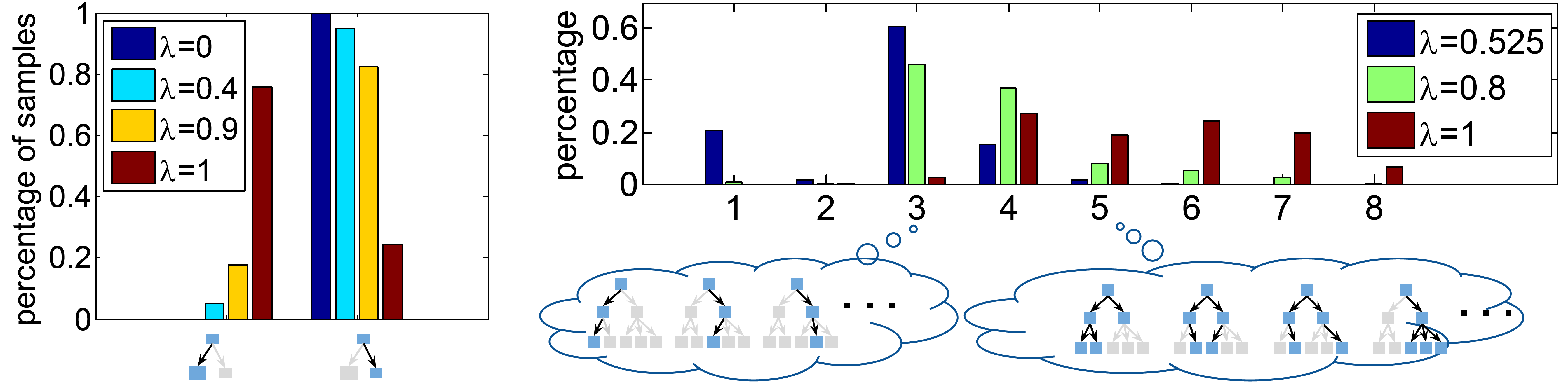}
\caption{Distribution of examples going through different execution paths. Skipped nodes
are in grey. The
  hyperparameter $\lambda$ controls the trade-off between accuracy and efficiency. A bigger
  $\lambda$ values accuracy more. \emph{Left}: for the high-low
  capacity D$^2$NN. \emph{Right}: for the hierarchical D$^2$NN. The X-axis
  is the number of nodes activated.  }
  \label{fig:hist_3layer}
\end{figure*}

As we can see, the D$^2$NN achieves a trade-off curve close to the
upperbound: there are points on the curve that are as fast as the low-capacity node and as
accurate as the high-capacity node.  Fig.~\ref{fig:hist_3layer}(left) plots the distribution of examples going through different
execution paths. It shows that as $\lambda$ increases, accuracy becomes more important and
more examples go through the high-capacity node. These results suggest that our 
learning algorithm is effective for networks with a single control node. 

With inference efficiency improved, we also observe that for training, a D$^2$NN typically takes 2-4 times more iterations to converge than a DNN, depending on particular model capacities, configurations and trade-offs. 

\smallparagraph{Cascade D$^2$NN}
We next experiment with a more sophisticated design that we call a ``cascade
D$^2$NN'' (Fig.~\ref{fig:d2nn_4all}b). It is inspired by the standard cascade design commonly used in computer vision. The intuition is
that many negative examples may be rejected early using simple features. The cascade D$^2$NN consists of seven regular
nodes (N1-N7) and three control nodes (Q1-Q3). N1-N7 form 4 cascade stages (i.e. 4 conventional DNNs, from small to large) of the cascade: N1+N2, N3+N4, N5+N6, N7. Each control node decides whether to
execute the next cascade stage or not. 

We evaluate the network on the same LFW-B face classification task using the same
evaluation protocol as in the high-low capacity
D$^2$NN. 
Fig.~\ref{fig:curve_4}b) plots the accuracy-cost tradeoff curve for the D$^2$NN.
Also included are the accuracy-cost curve (``static NNs'') achieved by the four conventional DNNs as baselines, each trained with a 
cross-entropy loss. We can see that the cascade D$^2$NN can achieve a close to optimal
trade-off, reducing computation significantly with negligible loss of accuracy. 
In addition, we can see that our D$^2$NN curve outperforms the trade-off curve achieved by varying the design and capacity of static conventional networks. 
This result demonstrates that our algorithm is successful for
jointly training multiple control nodes. 

For a cascade, wall time of inference is often an important consideration. 
Thus we also measure the inference wall time (excluding data loading with 5 runs) in this Cascade D$^2$NN. We find that a 82\% wall-time cost corresponds to a 53\% number-of-multiplication cost; and a 95\% corresponds to a 70\%. Defining reward directly using wall time can further reduce the gap.

\begin{figure*} \centering
\includegraphics[width=0.95\linewidth]{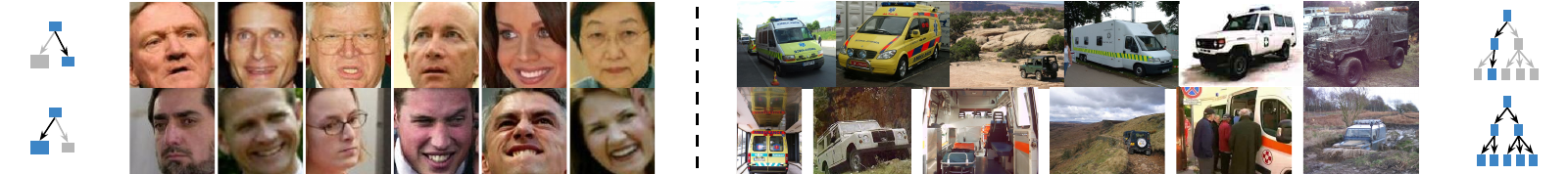}
   \caption{Examples with different paths in a high-low D$^2$NN (left) and a hierarchical D$^2$NN (right).}
\label{fig:vis_highlow}
\end{figure*}

\begin{figure} \centering
\includegraphics[width=.3\textwidth]{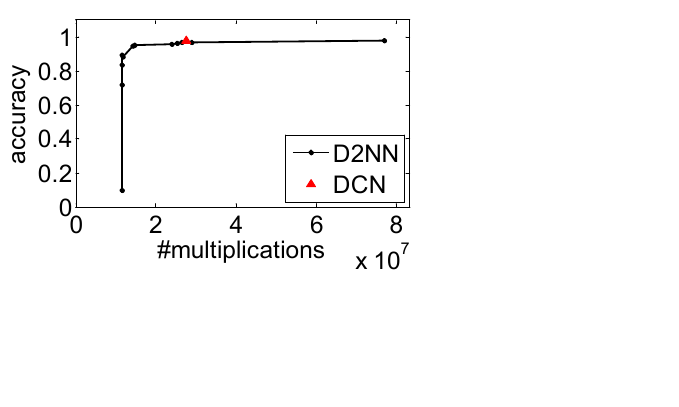}
\caption{Accuracy-cost curve for a chain D$^2$NN on the CMNIST task compared to DCN~\cite{DBLP:conf/icml/AlmahairiBCZLC16}. }
\label{fig:cmnist_curve}
\end{figure}

\smallparagraph{Chain D$^2$NN}
Our third design is  a ``Chain D$^2$NN'' (Fig.~\ref{fig:d2nn_4all}c).
The network is shaped as a chain, where each link consists of a control node selecting between
two (or more) regular nodes. In other words, we perform a sequence of vector-to-vector transforms; for each transform
we choose between several subnetworks. 
One scenario that we can use this D$^2$NN is that the configuration of a conventional DNN (e.g. number of layers, filter sizes) cannot be fully decided. Also, it can simulate shortcuts between any two layers by using an identity function as one of the transforms.
This chain D$^2$NN is qualitatively
different from other D$^2$NNs with a tree-shaped \emph{data graph} because it allows
two divergent \emph{data paths} to merge again. That is, the number of possible execution paths can be exponential to the number of nodes.

In Fig.~\ref{fig:d2nn_4all}c),  the first link is that Q1 chooses between a
low-capacity N2 and a high-capacity N3. If one of them is chosen, the other
will output a default value zero. The node N4 adds the outputs of N2 and N3 together. 
Fig.~\ref{fig:curve_4}c) plots the accuracy-cost curve on the LFW-B 
task. The two baselines are: a conventional DNN with the lowest capacity path (N1-N2-N5-N8-N10), and a conventional DNN with the highest capacity path (N1-N3-N6-N9-N10). 
The cost is measured as the number of multiplications, normalized by the cost of the high-capacity baseline. 

Fig.~\ref{fig:curve_4}c) shows that the chain D$^2$NN achieves a trade-off
curve close to optimal and can speed up computation significantly with  little
accuracy loss.  This shows that our learning algorithm is effective for a
D$^2$NN whose data graph is a general DAG instead of a tree. 

\smallparagraph{Hierarchical D$^2$NN}
In this experiment we design a D$^2$NN for hierarchical multiclass
classification. The idea is to first classify images to coarse categories and then to
fine categories. This idea has been explored by numerous prior
works~\cite{liu2013probabilistic,bengio2010label,deng2011fast}, but here we show that
the same idea can be implemented via a D$^2$NN trained end to end.

We use ILSVRC-10, a subset of the ILSVRC-65~\cite{deng2012hedging}. In
ILSVRC-10, 10 classes are organized into a 3-layer hierarchy: 2 superclasses,
5 coarse classes and 10 leaf classes.
Each class has 500 training images, 50 validation images, and 150 test images. 
As in Fig.~\ref{fig:d2nn_4all}d), the hierarchy 
in this D$^2$NN mirrors the semantic hierarchy in ILSVRC-10.  An image
first goes through the root N1. Then Q1 decides whether to descend
the left branch (N2 and its children), and Q2 decides whether
to descend the right branch (N3 and its children). The leaf nodes N4-N8 are each responsible 
for classifying two fine-grained leaf classes. 
It is important to note that an input image can go down parallel paths in the
hierarchy, e.g. descending both the left branch and the right
branch, because Q1 and Q2 make separate decisions. This ``multi-threading'' allows the network to avoid
committing to a single path prematurely if an input image is ambiguous.

Fig.~\ref{fig:curve_4}d) plots the accuracy-cost curve of our hierarchical
D$^2$NN. The accuracy is measured as the proportion of correctly classified test
examples. The cost is measured as the number of multiplications, normalized by the cost of
a conventional DNN consisting only of the regular nodes (denoted as NN in the figure).
We can see that
the hierarchical D$^2$NN can match the accuracy of the full network with about half of the computational
cost.

Fig.~\ref{fig:hist_3layer}(right) plots for the hierarchical D$^2$NN the distribution of examples going through
execution sequences with different numbers of nodes activated. Due to the parallelism of
D$^2$NN, there can be many different execution sequences. We also see that as $\lambda$
increases, accuracy is given more weight and more nodes are activated. 

\smallparagraph{Comparison with Dynamic Capacity Networks}
In this experiment we empirically compare our approach to closely related prior work. 
Here we compare D$^2$NNs with Dynamic Capacity Networks
(DCN)~\cite{DBLP:conf/icml/AlmahairiBCZLC16}, for which efficency measurement is
the absolute number of multiplications. Given an image, a DCN applies an additional high capacity subnetwork to a set of image patches, selected using a
hand-designed saliency based policy. The idea is that more intensive processing is only
necessary for certain image regions. 

To compare, we evaluate with the same multiclass classification task on the Cluttered
MNIST~\cite{mnih2014recurrent}, which consists of MNIST digits randomly placed on a
background cluttered with fragments of other digits. We train a chain D$^2$NN of length 4
, which implements the same idea of choosing a high-capacity alternative subnetwork for
certain inputs. Fig.~\ref{fig:cmnist_curve} plots the accuracy-cost curve of our D$^2$NN
as well as the accuracy-cost point achieved by the 
DCN in ~\cite{DBLP:conf/icml/AlmahairiBCZLC16}---an accuracy of $0.9861$ and
and a cost of $2.77\times10^7$. The closest point on our curve is an
slightly lower accuracy of $0.9698$ but slightly better efficiency (a cost of $2.66\times10^7$). Note
that although our accuracy of $0.9698$ is lower, it compares favorably to those of other state-of-the-art methods
such as DRAW~\cite{gregor2015draw}: $0.9664$ and RAM~\cite{mnih2014recurrent}: $0.9189$. 

\smallparagraph{Visualization of Examples in Different Paths}
In Fig.~\ref{fig:vis_highlow} (left), we show face examples in the high-low D$^2$NN for $\lambda$=0.4. Examples in low-capacity path are generally easier (e.g. more frontal) than examples in high-capacity path. 
In Fig.~\ref{fig:vis_highlow} (right), we show car examples in the hierarchical D$^2$NN with 1) a single path executed and 2) the full graph executed (for $\lambda$=1). They match our intuition that examples with a single path executed should be easier (e.g. less occlusion) to classify than examples with the full graph executed.

\smallparagraph{CIFAR-10 Results}
We train a Cascade D$^2$NN on CIFAR-10 where the corresponded DNN baseline is the ResNet-110. We initialize this D$^2$NN with pre-trained ResNet-110 weights, apply cross-entropy losses on regular nodes, and tune the mixed-loss weight as explained in Sec. 4. 
We see a 30\% reduction of cost with a 2\% loss (relative) on accuracy, and a 62\% reduction of cost with a 7\% loss (relative) on accuracy. 
The D$^2$NN's ability to improve efficiency relies on the assumption that not all inputs require the same amount of computation. In CIFAR-10, all images are low resolution (32 $\times$ 32), and it is likely that few images are significantly easier to classify than others. As a result, the efficiency improvement is modest compared to other datasets.

\section{Conclusion}
We have introduced Dynamic Deep Neural Networks (D$^2$NN), a new type of feed-forward deep neural
networks that allow selective execution. Extensive experiments have demonstrated that
D$^2$NNs are flexible and effective for optimizing accuracy-efficiency trade-offs. 

\section{Acknowledgments}
This work is partially supported by the National Science Foundation under
Grant No. 1539011 and gifts from Intel.

\section*{Appendix}
\appendix
\renewcommand{\appendixname}{Appendix~\Alph{section}}
\section{Implementation Details}

  We implement the D$^2$NN framework in Torch~\cite{torch}. Torch
  already provides implementations of conventional neural network modules
  (nodes). So a user can specify the subnetwork architecture inside a
  control node or a regular node using existing Torch functionalities. Our framework then
  handles the communication between the user-defined nodes in the forward and backward
  pass. 

   To handle parallel paths, default-valued nodes and nodes with multiple data parents,
   we need to keep track of an example's execution status (which nodes are activated by this example)
   and output status (which nodes have output for this example). An example's output status is different from its execution status if some nodes are not
   activated but have default values.
   For runtime efficiency, we implement the tracking of examples at
   the mini-batch level. That is, we perform forward and backward passes for
   a mini-batch of examples as a regular DNN does. Each mini-batch consists of
   several mini-bags of images.
  
   We describe the implementation of D$^2$NN learning procedure as two steps.
   First, the preprocessing step:
   When a user-defined D$^2$NN model is fed into our framework, we first
   perform a breadth-first search to get the DAG orders of nodes
   while performing structure error checks, contructing data and control
   relationships between nodes and calculating the cost (number of multiplications) of each node.
  
   After the preprocessing, the training step is similar to a regular DNN: a forward pass and a backward pass.
   All nodes are visited according to a topological ordering in a forward pass and the reverse ordering in a backward pass.
 
   For each function node, the forward pass has three steps: fetch inputs, forward inside the node, and
   send data or control signals to children nodes. When dealing with multiple data
   inputs and multiple control signals, the D$^2$NN will filter examples with more than one null inputs or
   all negative control signals. When a default value has been set for a node, all examples have to send out data. If the node is not activated for a particular example, the output will take the default value.
   A backward pass has similar logic: fetch gradients from children, perform the backward pass inside and
   send out gradients to parents. It is worth noting that when a default value is used in a node, the gradients
   can be blocked by this node because it is not actually executed.
  
\section{ILSVRC-10 Semantic Hierarchy}

\begin{figure*}[t] \centering
\includegraphics[width=0.8\textwidth]{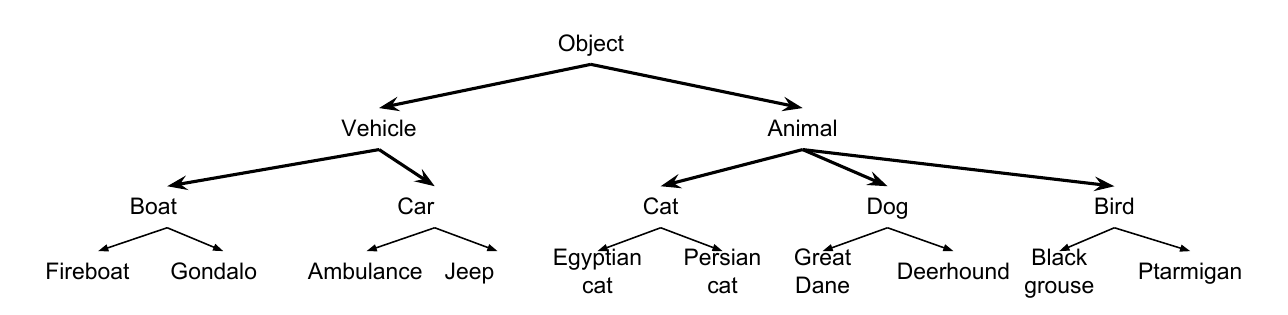}
\caption{The semantic class hierarchy of the ILSVRC-10 dataset.}
\label{fig:class}
\end{figure*}

The ILSVRC-10 dataset is a subset of the ILSVRC-65 dataset~\cite{deng2012hedging}. In our
ILSVRC-10, there are 10 classes organized into a 3-layer hierarchy: 2 superclasses,
5 coarse classes and 10 leaf classes as in Fig~\ref{fig:class}.
Each class has 500 training images, 50 validation images, and 150 test images. 

\section{Configurations}  

\smallparagraph{High-Low Capacity D$^2$NN}
The high-low capacity D$^2$NN consists of a single control node (Q) and three regular nodes
(N1,N2,N3) as illustrated in Fig.~\ref{fig:d2nn_4all}a). 

\begin{itemize}
\item Node N1: a convolutional layer with a 3$\times$3 filter size, 8 filters and a stride of 2,
  followed by a 3$\times$3 max-pooling layer with a stride of 2.
\item Node N2: a convolutional layer with a 3$\times$3 filter size and 16 filters,
  followed by a 3$\times$3 max-pooling layer with a stride of 2. The output is reshaped
  and fed into a fully connected layer with 512 neurons followed by another fully connected
  layer with the 2-class output.
\item Node N3: three 3$\times$3 max-pooling layers, each with a stride of 2, followed
  by two fully connected layers with 32 neurons and the 2-class output.
\item Node Q1: a convolutional layer with a 3$\times$3 filter size and 2 filters,
  followed by a 3$\times$3 max-pooling layer with a stride of 2. The output is reshaped
  and fed into a fully connected layer with 128 neurons followed by another fully connected
  layer with the 2-action output.
\end{itemize}

\smallparagraph{Cascade D$^2$NN}
The cascade D$^2$NN consists of a sequence of four regular
nodes (N1 to N7) and three control nodes (Q1-Q3) as in Fig.~\ref{fig:d2nn_4all}b).

\begin{itemize}
\item Node N1: a convolutional layer with a 3$\times$3 filter size, 2 filters and a stride of 2,
  followed by a 3$\times$3 max-pooling layer with a stride of 2.
\item Node N2: three 3$\times$3 max-pooling layers with strides of 2. The output is reshaped and 
  fed into a fully connected layer with the 2-class output.
\item Node N3: two convolutional layers with both 3$\times$3 filter sizes and 2, 8 filters respectively,
  each followed by a 3$\times$3 max-pooling layer with a stride of 2.
\item Node N4: two 3$\times$3 max-pooling layers with strides of 2. The output is reshaped and 
  fed into a fully connected layer with the 2-class output.
\item Node N5: two convolutional layers with both 3$\times$3 filter sizes and 4, 16 filters respectively,
  each followed by a 3$\times$3 max-pooling layer with a stride of 2.
\item Node N6: two 3$\times$3 max-pooling layers with strides of 2. The output is reshaped and 
  fed into a fully connected layer with the 2-class output.
\item Node N7: five convolutional layers with all 3$\times$3 filter sizes and 2, 8, 32, 32, 64 filters repectively,
  each followed by a 3$\times$3 max-pooling layer with a stride of 2 except for the third and fifth layer. The output is reshaped
  and fed into a fully connected layer with 512 neurons followed by another fully connected
  layer with the 2-class output.
\item Node Q1, Q2, Q3: the input is reshaped
  and fed into a fully connected layer with the 2-action output.
\end{itemize}

\smallparagraph{Chain D$^2$NN}
The Chain D$^2$NN is shaped as a chain, where each link consists of a control node selecting between
two regular nodes. In the experiments of LFW-B dataset, we use a 3-stage Chain D$^2$NN as in Fig.~\ref{fig:d2nn_4all}c).

\begin{itemize}
\item Node N1: a convolutional layer with a 3$\times$3 filter size, 2 filters and a stride of 2,
  followed by a 3$\times$3 max-pooling layer with a stride of 2.

\item Node N2: a convolutional layer with a 1$\times$1 filter size and 16 filters.
  
\item Node N3: a convolutional layer with a 3$\times$3 filter size and 16 filters.
  
\item Node N4: a 3$\times$3 max-pooling layer with a stride of 2.
  
\item Node N5: a convolutional layer with a 1$\times$1 filter size and 32 filters.
  
\item Node N6: two convolutional layers with both 3$\times$3 filter sizes and 32, 32 filters repectively.

\item Node N7: a 3$\times$3 max-pooling layer with a stride of 2.

\item Node N8: a convolutional layer with a 1$\times$1 filter size and 32 filters followed by a 3$\times$3 max-pooling layer with a stride of 2.
  The output is reshaped and fed into a fully connected layer with 256 neurons.

\item Node N9: a convolutional layer with a 3$\times$3 filter size and 64 filters.
  The output is reshaped and fed into a fully connected layer with 256 neurons.

\item Node N10: a fully connected layer with the 2-class output.
  
\item Node Q1: a convolutional layer with a 3$\times$3 filter size and 8 filters with a 3$\times$3 max-pooling layer with a stride of 2 before and
   a 3$\times$3 max-pooling layer with a stride of 2 after.
  The output is reshaped and fed into two fully connected layers with 64 neurons and the 2-action output respectively.

\item Node Q2: a 3$\times$3 max-pooling layer with a stride of 2 followed by a convolutional layer with a 3$\times$3 filter size and 4 filters.
  The output is reshaped and fed into two fully connected layers with 64 neurons and the 2-action output respectively.

\item Node Q3: a convolutional layer with a 3$\times$3 filter size and 2 filters.
  The output is reshaped and fed into two fully connected layers with 64 neurons and the 2-action output respectively.

\end{itemize}

\smallparagraph{Hierarchical D$^2$NN}
Fig.~\ref{fig:d2nn_4all}d) illustrates the design of our hierarchical D$^2$NN.

\begin{itemize}
\item Node N1: a convolutional layer with a 11$\times$11 filter size, 64 filters, a stride of 4 and a 2$\times$2 padding,
  followed by a 3$\times$3 max-pooling layer with a stride of 2.

\item Node N2 and N3: a convolutional layer with a 5$\times$5 filter size, 96 filters and a 2$\times$2 padding.
  
\item Node N4~N8: a 3$\times$3 max-pooling layer with a stride of 2 followed by three convolutional layers with 3$\times$3 filter sizes and
  160, 128, 128 filters respectively. The output is fed into a 3$\times$3 max-pooling layer with a stride of 2 and three fully connected layers
  with 2048 neurons, 2048 neurons and the 2 fine-class output respectively.
  
\item Node Q1 and Q2: two convolutional layers with 5$\times$5, 3$\times$3 filter sizes and 16, 32 filters respectively (the former has a 2$\times$2 padding), each
  followed by a 3$\times$3 max-pooling layer with a stride of 2.
  The output is reshaped and fed into three fully connected layers with 1024 neurons, 1024 neurons and the 2-action output respectively.

\item Node Q3~Q7: two convolutional layers with 5$\times$5, 3$\times$3 filter sizes and 16, 32 filters respectively (the former has a 2$\times$2 padding), each
  followed by a 3$\times$3 max-pooling layer with a stride of 2.
  The output is reshaped and fed into three fully connected layers with 1024 neurons, 1024 neurons and the 2-action output respectively.
\end{itemize}

\smallparagraph{Comparison with Dynamic Capacity Networks}
We train a chain D$^2$NN of length
4 similar to Fig.~\ref{fig:d2nn_4all}c).

\begin{itemize}
\item Node N1: a convolutional layer with a 3$\times$3 filter size and 24 filters.

\item Node N3: a convolutional layer with a 3$\times$3 filter size and 24 filters.
  
\item Node N4: a 2$\times$2 max-pooling layer with a stride of 2.
  
\item Node N6: a convolutional layer with a 3$\times$3 filter size and 24 filters.
  
\item Node N7: an identity layer which directly uses inputs as outputs.

\item Node N9: a convolutional layer with a 3$\times$3 filter size and 24 filters.

\item Node N10: a 2$\times$2 max-pooling layer with a stride of 2.

\item Node N12: a convolutional layer with a 3$\times$3 filter size and 24 filters.

\item Node N2, N5, N8, N11: an identity layer.
  
\item Node N13: a convolutional layer with a 4$\times$4 filter size, 96 filters, a stride of 2 and no padding,
  followed by a 11$\times$11 max-pooling layer.
  The output is reshaped and fed into a fully connected layer with the 10-class output.
 
\item Node Q1:  a convolutional layer with a 3$\times$3 filter size and 8 filters with two 2$\times$2 max-pooling layers with strides of 2 before
  and one 2$\times$2 max-pooling layer with a stride of 2 after.
  The output is reshaped and fed into two fully connected layers with 256 neurons and the 2-action output respectively.

\item Node Q2:  a convolutional layer with a 3$\times$3 filter size and 8 filters with a 2$\times$2 max-pooling layer with a stride of 2 before
  and a 2$\times$2 max-pooling layer with a stride of 2 after.
  The output is reshaped and fed into two fully connected layers with 256 neurons and the 2-action output respectively.

\item Node Q3:  a convolutional layer with a 3$\times$3 filter size and 8 filters with a 2$\times$2 max-pooling layer with a stride of 2 before
  and a 2$\times$2 max-pooling layer with a stride of 2 after.
  The output is reshaped and fed into two fully connected layers with 256 neurons and the 2-action output respectively.

\item Node Q4:  a convolutional layer with a 3$\times$3 filter size and 8 filters,
  followed by a 2$\times$2 max-pooling layer with a stride of 2.
  The output is reshaped and fed into two fully connected layers with 256 neurons and the 2-action output respectively.

\end{itemize}

For all 5 D$^2$NNs, all convolutional layers use 1$\times$1 padding and each is followed by a ReLU layer unless specified individually.
Each fully connected layer except the output layers is followed by a ReLU layer.

{\small \bibliographystyle{ieee} \bibliography{egbib} }

\end{document}